\documentclass[11pt]{article}
\usepackage[final]{acl}
\usepackage{microtype}

\usepackage{fontspec}
\newfontfamily\telugufont{NotoSansTelugu-Regular.ttf}
\newcommand{\texttelugu}[1]{{\telugufont #1}}

\usepackage{amsmath}
\usepackage{array}
\usepackage{multirow}
\usepackage{graphicx}
\usepackage{tabularx}
\usepackage{url}
\usepackage{xcolor}
\usepackage{float}
\usepackage{ragged2e}
\usepackage{inconsolata}

\title{Human-Centered Supervision for Sentiment Analysis in Telugu: A Systematic Inquiry Beyond Accuracy}
\author{
\textbf{Vallabhaneni Raj Kumar*}\textsuperscript{1}\!,
\textbf{Ashwin S*}\textsuperscript{1}\!,
\textbf{Supriya Manna*}\textsuperscript{2,$\ddagger$}\!,
\textbf{Niladri Sett}\textsuperscript{3,$\ddagger$,$\dagger$}\!,\\
\hspace{3pt}\textbf{Cheedella V S N M S Hema Harshitha}\textsuperscript{1}\!,
\textbf{Kurakula Harshitha}\textsuperscript{1}\!,
\textbf{Anand Kumar Sharma}\textsuperscript{1}\!,\\
\textbf{Basina Deepakraj\textsuperscript{1}}\!,
\textbf{Tanuj Sarkar}\textsuperscript{1}\!,
\textbf{Bondada Navaneeth Krishna}\textsuperscript{1}\!,
\textbf{Samanthapudi Shakeer}\textsuperscript{1}
\\[0.5em]
\textsuperscript{1}SRM University AP, India \hspace{.2em}
\textsuperscript{2}University of Maryland, College Park, USA \hspace{.2em}
\textsuperscript{3}GITAM University, Bangalore, India
\\[0.5em]
\centering \textsc{$\odot$ Project Page}: \url{https://huggingface.co/DSL-13-SRMAP}
}

\begin{document}

\maketitle

\begingroup
\setlength{\parindent}{0pt}
\renewcommand\thefootnote{}
\footnotetext{
\textsuperscript{*}Equal contribution; order determined randomly.

\textsuperscript{$\ddagger$}Most of the work was completed while the author was with SRM University AP, India.

\textsuperscript{$\dagger$}Corresponding author: settniladri@gmail.com
}
\endgroup
\setcounter{footnote}{0}

\begin{abstract}
Sentiment analysis for low-resource languages remains challenging in an era where interpretability, human alignment, and fairness are increasingly non-negotiable aspects of modern machine learning systems. These challenges stem both from the scarcity of annotated data and from the resulting difficulty of conducting reliable, human-interpretable analyses that go beyond predictive accuracy. Telugu, one of the primary Dravidian languages with over 96 million speakers, is not an exception. In this work, we first introduce TeSent, a large-scale Telugu sentiment classification dataset annotated with sentiment labels and human-selected rationales from multiple native speakers. This resource enables the study of rationale-based supervision for aligning models with human reasoning in this low-resource setting. We fine-tune five transformer-based models with and without rationale supervision and evaluate them on classification performance, explanation quality, and social bias. To facilitate controlled fairness evaluation, we additionally construct TeEEC, an evaluation corpus for Telugu sentiment analysis. Our results show that incorporating human rationales consistently improves alignment and often leads to holistic gains in predictive performance. We further provide extensive analysis of multi-facade explanation quality and fairness, offering insights into the broader effects of alignment-oriented supervision in resource-scarce language contexts.
\end{abstract}

\section{Introduction}

Language plays a central role in how societies think, communicate, and pass knowledge across generations. Telugu, which belongs to the Dravidian language family native to South India, has a long recorded history, with inscriptions dating back to around 575 CE and a well-established literary tradition by the 11th century. It is now recognized as one of India’s six classical languages and is spoken by more than 96 million people worldwide\footnote{\href{https://www.ethnologue.com/language/tel/}{https://www.ethnologue.com/language/tel/}}. In spite of this wide usage and cultural importance, Telugu has received relatively little attention in natural language processing and machine learning research. The limited availability of carefully curated and thoroughly evaluated datasets has slowed progress, and many language-specific characteristics of Telugu are still not well studied. In this work, we attempt to address this gap by presenting a benchmark for Telugu sentiment classification and by examining several research questions that are important for current machine learning approaches.

Sentiment annotation involves interpretation in addition to classification. While a sentiment label records an annotator’s final decision, it does not capture the reasoning behind that decision. As a result, benchmarks that rely only on final labels, which is common in prior work~\citep{pang2002thumbs,socher2013recursive,maas2011learning,dong2014adaptive,wang2010latent,rosenthal2017semeval,orbach2021yaso}, provide limited insight into human judgment. This limitation is particularly relevant for low-resource settings, where annotation quality and interpretability play a critical role. In this work, we move beyond label-only supervision by modeling human preference at two levels. We treat sentiment labels from multiple annotators as objective outcomes, and annotators’ rationales as subjective explanations for those outcomes. We further distinguish between primary and secondary rationales to reflect differences in salience during annotation. To our knowledge, this is the first Indic-language dataset that jointly models sentiment decisions and their associated justifications.

This human-centered data foundation allows us to study questions that have so far been difficult to explore for Telugu. Instead of treating sentiment analysis as a purely label-driven task, we examine how incorporating human understanding into supervision affects model behavior and reasoning in a low-resource setting. In particular, we focus on the role of subjective human preference, captured through annotator rationales, and study how such supervision influences learning, interpretability, and downstream behavior. Our analysis is guided by three research questions: \textbf{(RQ1)} \textit{\textbf{h}ow the inclusion of human rationales alongside sentiment labels changes model behavior compared to label-only supervision}; \textbf{(RQ2)} \textit{\textbf{h}ow human-centered supervision affects the alignment between model explanations and human reasoning}; and \textbf{(RQ3)} \textit{\textbf{h}ow alignment-oriented design choices interact with broader model properties, including social bias and fairness, in a low-resource language context.}

To the best of our knowledge, high-quality Telugu sentiment resources remain scarce even within the broader Indic language landscape. Among existing efforts, only a few Telugu sentiment corpora are worth mentioning~\citep{mukku2017actsa,marreddy2022resource}, and most exhibit one or more limitations: small scale (often fewer than 10{,}000 instances~\citep{mukku2017actsa,chakravarthi2020sentiment}); insufficient documentation of annotation protocols~\citep{marreddy2022resource,kulkarni2021l3cubemahasent,akhtar2016hybrid}; weak annotation setups, commonly involving fewer than three annotators per instance~\citep{doddapaneni2023towards,kulkarni2021l3cubemahasent,patwa2020semeval,chakravarthi2020sentiment,mukku2017actsa}; narrow domain coverage~\citep{chakravarthi2020sentiment,chakravarthi2020corpus,mukku2017actsa,kulkarni2021l3cubemahasent,patwa2020semeval}; and reliance on translated rather than native Telugu text~\citep{doddapaneni2023towards}. These constraints limit not only dataset quality but also the study of human reasoning, alignment, and interpretability in Telugu. To address these gaps, we introduce the following components:
\vspace{-1em}
\begin{itemize}
  \item A human-annotated Telugu dataset of 21,119 sentence-level instances for three-class sentiment classification\footnote{We focus on sentence-level sentiment classification, assigning each sentence a polarity from \textit{positive}, \textit{negative}, or \textit{neutral}. Unless stated otherwise, sentiment classification refers to this setting.}. Each instance includes a sentiment label (objective preference) and textual rationales capturing subjective human reasoning.
\vspace{-.5em}
  \item Five fine-tuned pre-trained SOTA models trained with and without human rationales, enabling controlled analysis of the impact of human-centered supervision.
  \vspace{-.5em}
  \item \textbf{TeEEC} (\textbf{Te}lugu \textbf{E}quity \textbf{E}valuation \textbf{C}orpus), a fairness evaluation corpus for sentiment and emotion-related NLP tasks in Telugu, with benchmarks for gender and religion bias.
  \vspace{-.5em}
  \item A human-rationale–grounded explainability framework supporting plausibility and faithfulness evaluation using six widely used post-hoc explanation methods.
\end{itemize}
\vspace{-.5em}

We collect user comments from YouTube, news websites, and blogs across approximately 20 domains and curate them using a preprocessing pipeline that removes code-mixed content, duplicates, and non-Telugu text. Using a custom annotation setup, each sentence is labeled by three native Telugu speakers drawn from a gender-balanced set of 95 participants, with quality control and post-annotation validation in place. Our experiments indicate that incorporating human rationales improves sentiment classification performance and produces explanations better aligned with human reasoning.

We firmly believe our work will serve as a foundational stepping stone for inclusive, interpretable, and fair NLP research in Telugu and inspire similar efforts in other underrepresented languages.

\section{Related Works}
\label{RW}
Sentiment classification has been extensively studied in English, beginning with product-review datasets~\citep{pang2002thumbs,maas2011learning} and later extending to large-scale social media benchmarks that address informal, noisy text and multi-class sentiment settings~\citep{socher2013recursive,dong2014adaptive,wang2010latent,rosenthal2017semeval,orbach2021yaso}. In contrast, sentiment analysis for Indic languages remains severely under-resourced. Existing Indic datasets span a limited number of languages and often suffer from small-scale, narrow domain coverage, weak or undocumented annotation protocols, or reliance on translated rather than native text~\citep{akhtar2016hybrid,chakravarthi2020sentiment,patwa2020semeval,doddapaneni2023towards}. Telugu, despite its large speaker base, is particularly underrepresented, with only a small number of sentiment corpora available~\citep{mukku2017actsa,marreddy2022resource}. 

More broadly, existing Indic sentiment benchmarks do not support systematic investigation of human-centered supervision, explainability, or fairness. In English, text classification benchmarks that include human annotated rationales--including those for sentiment analysis--have primarily been used to improve model performance or to evaluate the plausibility of post-hoc explanation methods \citep{herrewijnen2024human}. More recently, Explanation-Guided Learning (EGL) \citep{gao2024going} has emerged as a research direction that incorporates human-annotated explanations directly into training and studies their effects on model behavior, explanation quality, and fairness, accountability, and transparency (FAccT) properties. However, to the best of our knowledge, no prior work has systematically explored EGL in Indic languages, or more broadly in low-resource language settings. Our work addresses this gap by introducing a large-scale, human-centered Telugu sentiment resource explicitly designed to support supervision beyond labels, explainability analysis, and fairness evaluation.

The paradigm of human alignment is multi-facade and broadly refers to techniques for steering model behavior toward human expectations \citep{christian2020alignment}. Prior human-centered supervision paradigms such as instruction tuning~\citep{wei2022finetuned} and preference learning~\citep{ouyang2022training} have predominantly been studied in the context of large language models performing text generation, where alignment is achieved through output-level or interaction-level feedback. In contrast, our work focuses on discriminative sentiment classification with pre-trained encoder-based models, incorporating annotator-provided rationales alongside sentiment labels as input-grounded supervision. This setting enables a systematic examination of how human-centered supervision influences classification behavior, the alignment of model explanations with human reasoning, and broader properties such as bias and fairness in a low-resource language.

A more detailed discussion of related benchmarks, explainability, and fairness literature is provided in Appendix~\ref{DRS}.

\section{Data Collection \& Preprocessing} \label{DCP}

To create a diverse corpus, we collected data from multiple sources: YouTube comments, comments from various Telugu blogs, news websites (\textit{\href{https://www.eenadu.net/}{Ennadu}}, \textit{\href{https://www.andhrajyothy.com/}{Andhra Jyothi}}, \textit{\href{https://telugu.way2news.com/}{Way2News (Telugu)}}, and \textit{\href{https://www.sakshi.com/}{Sakshi}}). The final corpus comprises 53.70\% YouTube comments, 28.20\% blog comments, and 18.10\% news headlines.

For YouTube, we curated a set of standard topics (e.g. Politics, Sports, etc.) and search phrases per topic (details provided in Appendix \ref{AC}). Using the \textit{YouTube API library} (version~3), we retrieved up to 50 top videos for each search phrase. We then extracted all comments that included at least one Telugu word or phrase using \textit{Langdetect} at 90\% confidence threshold. The whole extraction process took place between September 10th and 20th, 2024.

Regarding Telugu blogs, we initially sourced data from the website `\href{https://telugublogworld.blogspot.com/?m=0}{telugublogworld}', which aggregates a list of popular Telugu blogs. We used a custom script with \textit{BeautifulSoup} library (version~4), a Python library to recursively navigate through the hyperlinks from the webpages to verify whether there were active blog sites. From each available website, for all the blogs, we scraped as many comments as possible. This whole process took place between September 5th and 10th, 2024, and included content from active Telugu blogging platforms like: 
\href{https://avsfilm.blogspot.com/}{Avs film},
\href{https://malakpetrowdy.blogspot.com/}{malakpet rowdy},
\href{https://manchupallakee.blogspot.com/}{manchupallakee},
\href{https://ongoluseenu.blogspot.com/}{ongoluseenu},
\href{https://ekalingam.blogspot.com/}{ekalingam},
\href{https://sarath-kaalam.blogspot.com/}{sarath-kaalam},
\href{https://apmediakaburlu.blogspot.com/}{apmediakaburulu},
\href{https://kandishankaraiah.blogspot.com/}{kandishankaraiah},
\href{https://andhraamrutham.blogspot.com/}{andhraamrutham}, and
\href{https://indrachaapam.blogspot.com/}{indrachaapam}. We opted for this approach due to the often limited visibility (e.g., \href{https://neanoo-naakavitwam.blogspot.com/}{neanoo-naakavitwam}, \href{https://tekumalla-venkatappaiah.blogspot.com/}{tekumalla-venkatappaiah}), inactivity (e.g., \href{http://jalleda.com}{Jalleda}), restricted access (e.g., \href{https://telanganayasa.blogspot.com}{Telanganayasa}), and maintenance issues (e.g., \href{https://teluguwebmedia.com/}{TeluguWebMedia}, \href{https://telugubloggers.com/}{TeluguBloggers}, \href{https://malica.org}{Malica}) faced by many Telugu blogs, unlike English blogging platforms.

For news websites, using \textit{BeautifulSoup} (version~4), we scraped article headlines from the aforesaid news websites over one week, spanning 25th September to 2nd October. Note, this refers to the scraping period, not the publication dates of the articles\footnote{Along with this, we previously scraped Facebook and Instagram comments; however, due to ambiguities in Meta’s Terms of Service, these were ultimately excluded from the final version of TeSent.}.

The raw corpus was subsequently subjected to a comprehensive preprocessing pipeline involving language filtering, deduplication, and anonymization. For clarity and reproducibility, the complete preprocessing procedure is described in Appendix~\ref{App:Preprocessing}. In our annotation process, if an annotator finds a sentence grammatically incorrect, incomplete, off-topic, or uninterpretable, they can flag it as ``Mark as Bad Sentences." (\textit{invalid flag}) on the fly. Any sentence with at least one \textit{invalid flag} is marked \textit{invalid} and excluded from the dataset. After preprocessing, we first run an internal pilot study to hands-on the annotators with the annotation tasks with randomly selected 607 sentences, out of which 482 were valid, and from the annotated corpus of 23,986 sentences, 21,743 were found to be valid. This brought the total number of valid annotated sentences to 22,225. The description of our Annotation software used for this work is at Appendix \ref{App:Software}.

\section{Annotation Process}
\subsection{Overview}

The annotation process was designed to capture both sentiment polarity and fine-grained human rationales in Telugu text. A set of trained native Telugu speakers annotated each sentence using a structured, two-layer framework that balances annotation quality with cognitive efficiency. Details regarding annotator recruitment and our custom annotation software for doing it are provided in Appendix~\ref{App:Recruitment} and Appendix~\ref{App:Software}, respectively.

\subsection{Annotation Framework}
In contrast to prior efforts, we adopt a distinct methodological framework for annotation. To support this, we also developed custom in-house annotation software tailored to our specific design principles, as detailed in Appendix~\ref{App:Software}. Each annotator annotates every sentence following two layers of annotation. First, annotators are instructed to assign a primary sentiment label. Following this, they are asked to highlight specific words or phrases in the sentence that serve as human rationales for their chosen label. To capture subtler sentiment dynamics, annotators are then given the option to provide a secondary sentiment label if they believe one reasonably coexists with the primary annotation. However, for brevity and to reduce cognitive load, we do not collect rationales for this secondary label.

Incorporating an optional secondary sentiment label alongside the primary one allows us to better reflect the ambiguity and subjectivity that often arise in natural language, particularly in sentiment-laden content. Furthermore, we make rationale selection optional for sentences labelled as \textit{neutral}, acknowledging that such sentences may not contain explicit sentiment-bearing expressions. In contrast, for sentences labelled as \textit{positive} or \textit{negative}, rationale selection is mandatory to ensure the annotation captures the textual cues driving the sentiment decision.

\subsection{Finalizing the Label}\label{fin-label}
To finalize the sentiment label for each sentence, we begin by discarding annotations where the primary and secondary sentiments are direct contradictions--specifically, cases where the primary label is \textit{positive}, and the secondary is \textit{negative}, or vice versa (we denote this as `<+,-> pair'). Such contradictions may indicate either annotator inconsistency or that the sentence was noisy or ambiguous to them, and including these could compromise label quality. Following this filtering step, we apply a majority voting scheme over the primary sentiment labels provided by all annotators. If a clear majority emerges from that, we assign that label as the final one. In cases where there is no majority, i.e., when annotators choose all three different labels (positive, negative, and neutral), we then incorporate the available secondary labels and re-run the majority vote. However, in such cases, we discovered that although secondary sentiments help resolve the \textit{no majority} problem, the final label still could be confusing. Since such cases were rare in our dataset, we did not perform additional manual intervention and discarded the small split containing $\le 3.2\%$ of the sentences from the final dataset.

Out of 22,225 total valid and annotated sentences, 19,755 had only primary annotations from all annotators, of which 19,087 had a clear majority winner and were retained. The remaining 2,470 sentences included at least one annotation with both a primary and secondary label; after removing 58 due to a technical glitch, 2,412 remained. From these, annotations with <+, -> pairs were removed where present, and majority voting over the priority labels yielded 2,115 sentences. This resulted in a corpus of 21,202 sentences. Finally, Telugu-speaking authors manually inspected the entire annotated dataset and, at their discretion, removed an additional 83 malformed or duplicate sentences, leading to a final total of 21,119.  Overall, we retained over 95\% of high-quality sentences with the final label from the set of valid sentences. Lastly, with input from a senior Telugu-speaking co-author and using the off-the-shelf \href{https://huggingface.co/ai4bharat/IndicNER}{IndicNER} model, we performed named entity recognition and did not identify any offensive sentences to the best of our understanding.

We provide an overview of the dataset’s lexical and genre characteristics, including word-count statistics and genre distribution, in section ~\ref{stat}.

\section{Dataset Statistics} \label{stat}

\begin{figure}[ht]
    \centering
    \includegraphics[width=0.7\linewidth]{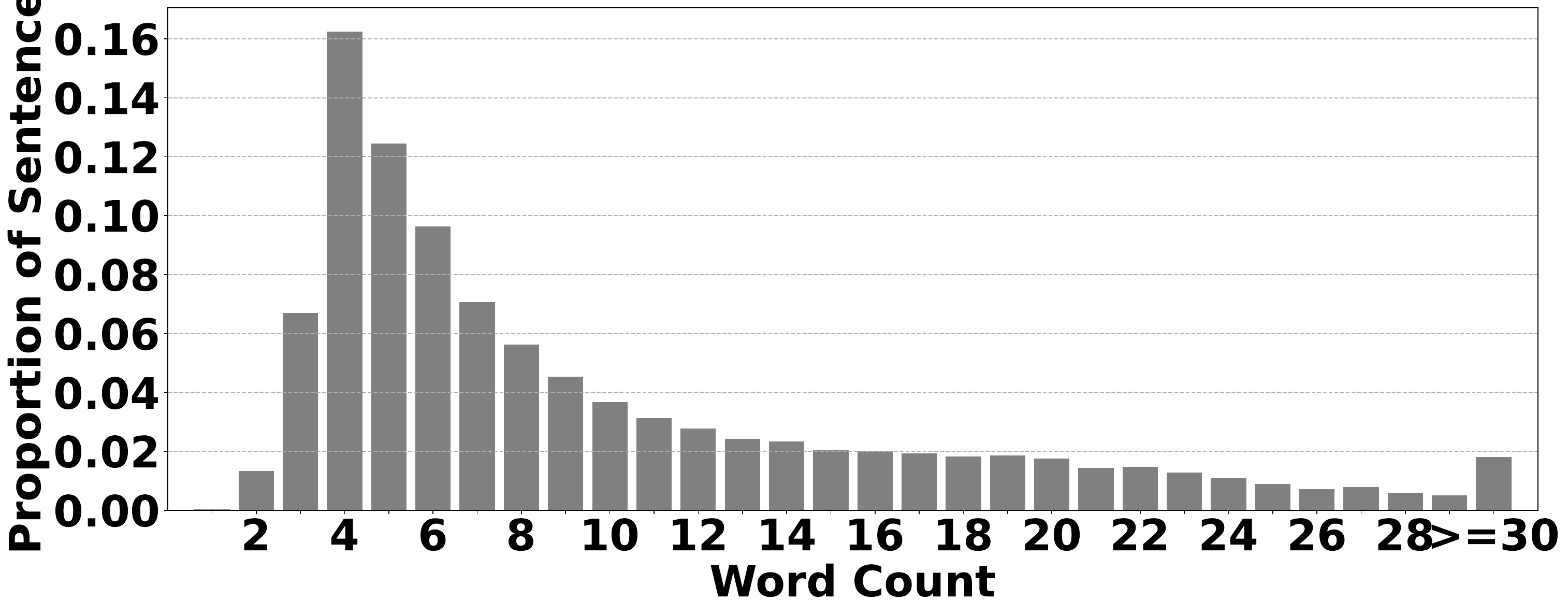}
     \caption{Word Count Frequency for TeSent}
    \label{fig:word_length}

\end{figure}

We refer the readers to Figure \ref{fig:word_length} for an overview of word count frequency in TeSent. 

\begin{figure}[ht]
    \centering
    \includegraphics[width=0.6\linewidth]{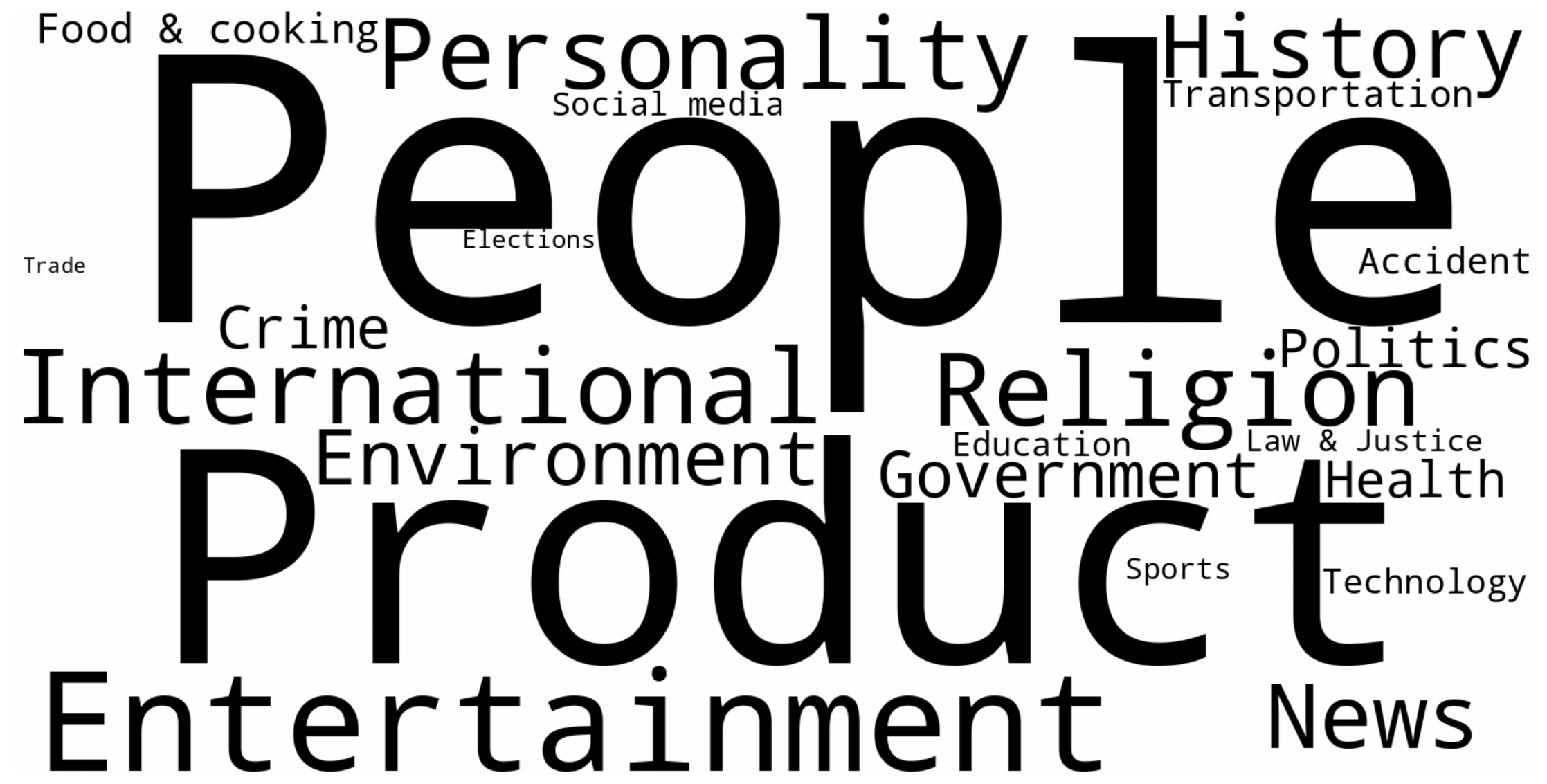}
    \caption{Category Wordcloud for TeSent}
    \label{category_frequency_wordcloud}
\end{figure}

To categorise the genre of the final dataset, we have translated all sentences into English and used the Hugging Face model \texttt{\href{https://huggingface.co/facebook/bart-large-mnli}{facebook/bart-large-mnli}} in single-label mode to assign the most relevant category to each sentence, as shown in Figure \ref{category_frequency_wordcloud}.

\begin{figure}
    \centering
    \includegraphics[width=.6\linewidth]{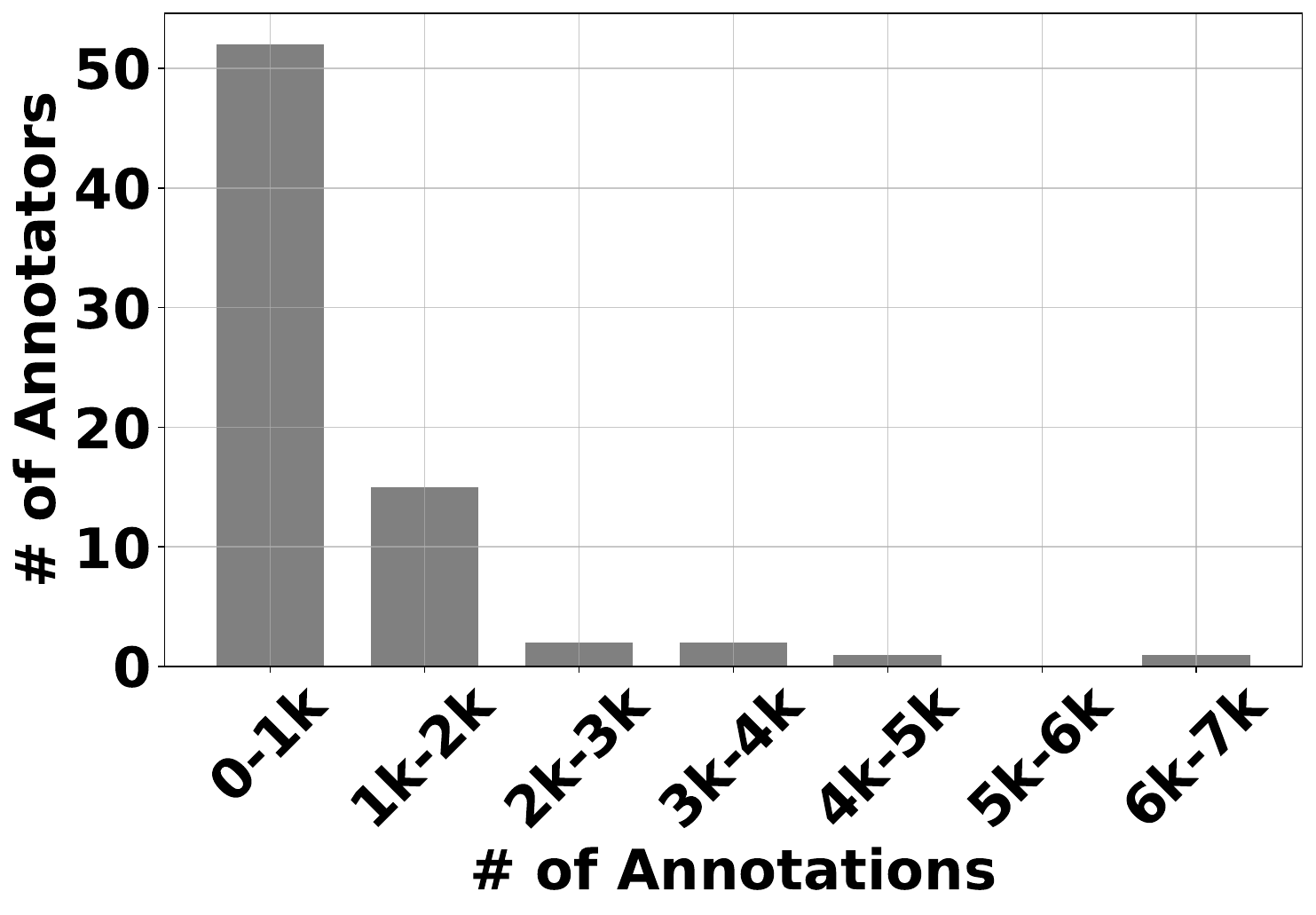}
    \caption{Annotation Distribution}
    \label{annotator_vs_annotations}
    \end{figure}

Distribution of annotation contributions across annotators in the TeSent dataset, showing the number of instances annotated per annotator is given in Figure~\ref{annotator_vs_annotations}.

We summarize the annotation procedure and dataset statistics in Table \ref{AnnonCard}.

\subsection{Annotation Statistics and Interpretation}
Our final dataset consists of 45.42\% neutral, 26.84\% positive, and 27.74\% negative instances, annotated by a large set of annotators with varying levels of contribution; additional details are provided in Appendix~\ref{stat}. Inter-annotator agreement was measured using Krippendorff’s Alpha ($\alpha$) and Fleiss’s Kappa ($\kappa$) \citep{gwet2014handbook}. For the three-class sentiment setting, we obtain $\alpha = 0.4706$ and $\kappa = 0.4705$.

Interpreting agreement coefficients in multi-class settings is less straightforward than in binary tasks, and commonly used qualitative labels (e.g., “moderate” or “strong”) are primarily derived from binary or near-binary scenarios \citep{mchugh2012interrater}. For this reason, we avoid assigning threshold-based qualitative interpretations to these values. Notably, the observed agreement is consistent with prior methodologically comparable studies; for example, \citet{mathew2021hatexplain} report a Krippendorff’s $\alpha$ of 0.46 for a three-class social-media annotation task with inherent ambiguity, which also incorporates human-annotated rationales alongside labels.

\begin{figure}
    \centering
    \includegraphics[width=0.25\textwidth]{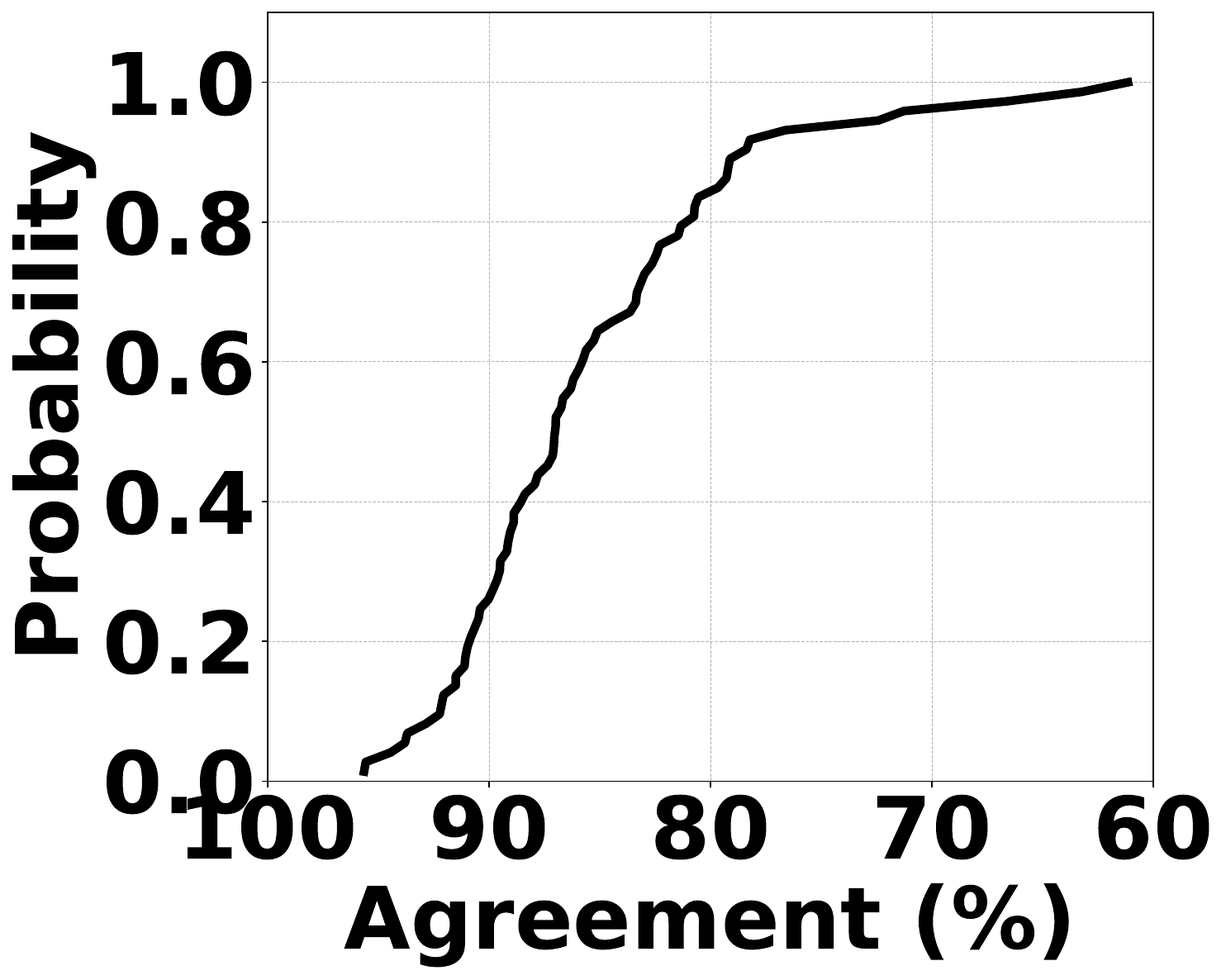}
    \caption{Distribution of Agreement Levels}
    \label{annotator_match_percentage}
\end{figure}

To provide a more interpretable view of annotator consistency, as suggested by prior work emphasizing complementary agreement analyses beyond summary coefficients~\citep{mchugh2012interrater}, we additionally report the cumulative distribution of pairwise agreement among annotators (Figure~\ref{annotator_match_percentage}). The distribution shows that approximately 80\% of annotators agree with the remaining annotators on more than 82\% of instances, and nearly all annotators exceed a 75\% agreement level, indicating that disagreement is concentrated in a relatively small subset of instances.

Finally, when collapsing the task to a binary sentiment setting by excluding neutral instances, agreement increases substantially to $\alpha = 0.8928$, indicating strong consistency when sentiment polarity is unambiguous. Taken together, the binary agreement, the agreement distribution in the three-class setting, and consistency with prior work support the reliability of the annotation process for capturing nuanced sentiment distinctions in Telugu.

\begin{table*}[t]
\centering
\small
\setlength{\tabcolsep}{4pt}
\begin{tabularx}{\textwidth}{p{2.1cm}|p{2.2cm}|p{2.2cm}|X|p{2.9cm}|p{2.3cm}}
\textbf{Classes} & \textbf{Setup} & \textbf{Annotator} & \textbf{Finalizing Label} & \textbf{Dataset} & \textbf{TTS} \\
\hline
$\bullet$Positive (\textbf{P})\newline $\bullet$Negative (\textbf{N})\newline $\bullet$Neutral (\textbf{Nt})
& $\bullet$Primary label \newline $\bullet$Secondary label (opt.) \newline $\bullet$Rationale
& $\bullet$74 annotators \newline $\bullet$3 per sentence \newline
& $\bullet$Majority voting \newline $\bullet$Invalid and no-majority annotations removed
& $\bullet$21,119 sentences \newline $\bullet$\textbf{P}:26.84\%,\textbf{N}:27.74\% \newline,\textbf{Nt}:45.42\% \newline $\bullet$$\alpha, \kappa = 0.47$
& $\bullet$Train: 16,896 \newline $\bullet$Val: 2,110 \newline $\bullet$Test: 2,113 \\
\hline
\end{tabularx}

\caption{Annotation card.}
\label{AnnonCard}
\end{table*}

\section{Fairness Benchmarking}
\label{Fair_Bench}
Sentiment classification systems can perpetuate various societal biases, including gender, racial, and identity-based biases. The presence of such bias often leads to disparities in classification accuracy or sentiment polarity across different demographic groups. Kiritchenko et al.~\citep{kiritchenko2018examining} proposed the Equity Evaluation Corpus (EEC), an English dataset designed to evaluate age and racial bias across multiple NLP tasks related to sentiment and emotion. We extend and tune EEC accommodating niche morphological nuances in Telugu language to create \textbf{TeEEC}, Telugu Equity Evaluation Corpus, and propose a robust fairness evaluation framework for sentiment classification in Telugu, with a focus on gender and religion bias. Although our bias evaluation framework is specifically designed for sentiment classification, TeEEC can be applied to any NLP task involving sentiment or emotion.

\subsection{TeEEC}
\label{TeEEC}

We construct the Telugu Equity Evaluation Corpus (TeEEC) to evaluate demographic fairness in Telugu sentiment classification using a controlled, template-based counterfactual framework derived from Kiritchenko et al.~\citep{kiritchenko2018examining}. TeEEC is based on 11 sentence templates originally proposed in English, comprising seven templates with explicit emotion words and four sentiment-neutral templates without emotion words (e.g., “<person> feels <emotion word>” and “I saw <person> in the market”). In the former, the sentiment polarity is determined solely by the emotion word, while in the latter, any difference in prediction can be attributed only to the demographic reference. All templates are translated into Telugu while preserving their original structure and intent, and no new templates are introduced, ensuring consistency with prior work and avoiding additional sources of variation. For the emotion-bearing templates, we select emotion words from four affective categories-Anger, Fear, Joy, and Sadness-and choose ten commonly used Telugu emotion words from each category; the inserted emotion word determines the sentiment polarity of the sentence, with morphological realization chosen to ensure grammatical correctness.\footnote{We consider two morphological realizations of emotion words--a noun form and a corresponding adjectival form--and select the form required by each template; a manual analysis shows that Templates 3, 4, and 7 (as numbered in~\citep{kiritchenko2018examining}) license noun forms, while the remaining templates require adjectival forms.}

TeEEC evaluates fairness with respect to gender $\{\text{male, female}\}$ and religion $\{\text{Hindu, Muslim, Christian}\}$, instantiated through the \texttt{<person>} placeholder using names or noun phrases associated with the corresponding group. For gender evaluation, we curate 20 commonly used Telugu names for each gender and 10 noun phrases per group, where the noun phrases exhibit a one-to-one correspondence across genders and do not encode religion; for religion evaluation, names are grouped by religious affiliation, and noun phrases are not used. We do not include caste, age, occupation, or socio-economic factors, although these are relevant in the Indian context, because they cannot be incorporated into this counterfactual framework without introducing ambiguity or severely limiting coverage.\footnote{Caste is often not unambiguously identifiable from names and is frequently conflated with religion or region; for example, surnames such as \textit{Naidu} or \textit{Reddy} may correspond to different communities across regions or persist across religious conversion (e.g., \textit{John Reddy}, \textit{Mary Kumari}), making it infeasible to generate unambiguous counterfactual sentence pairs. Directly inserting caste labels would also yield very few usable counterfactuals; similar challenges arise for occupation and age groups.}
Moreover, since our goal is to evaluate bias specifically for sentiment classification, each template must include a sentiment identifier, which makes curating specialized templates for additional factors infeasible within the scope of this work. Manually curating or annotating sentences (including with LLMs) for bias evaluation is outside the scope of this study and is left for future work\footnote{Hindu and Christian names were randomly sampled from the curator-maintained reference site \href{https://www.behindthename.com/names/list}{Behind the Name}, and Muslim names from the analogous resource \href{https://quranicnames.com/}{QuranicNames}. Within the best of our capacity, we verified whether sampled names appear in the respective religious texts using publicly available web sources; however, such verification was not possible for all names.}.

For gender bias evaluation, we generate two disjoint sentence sets corresponding to male and female references, each containing all possible combinations of demographic-specific \texttt{<person>} realizations and emotion words across all emotion-bearing templates, along with sentences derived from sentiment-neutral templates, resulting in a total of 39{,}760 sentences evenly split between the two gender groups; the religion bias dataset is constructed following the same procedure using names only. TeEEC thus provides, to the best of our knowledge, the first large-scale equity evaluation corpus for Telugu covering the two most prominent and identifiable societal dimensions--gender and religion--under a controlled and reproducible setup.

\subsection{Evaluation Metrics}
\label{Eval_Met}
We evaluate bias in two distinct ways. First, we measure differences in predicted sentiment polarities across demographic groups using a counterfactual test~\citep{kiritchenko2018examining,goldfarb2023bias}, which we formalize as the Polarity Difference Score ($PDS$). Second, we assess fairness using equalized odds (EO)~\citep{hardt2016equality}, computed as the absolute differences in true positive rates and false positive rates across groups, denoted as $TPR_{diff}$ and $FPR_{diff}$. Details of the counterfactual test, PDS, and EO are provided in Appendix \ref{F_MET}.

\section{Explainability Benchmarking}
Commonly used post-hoc local explainers can be broadly categorized into two types: \textit{perturbation-based} and \textit{gradient-based} explainers \citep{arrieta2020explainable}. In this work, we consider two widely adopted perturbation-based, model-agnostic explainers: \textbf{LIME} \citep{ribeiro2016should} and \textbf{SHAP} \citep{lundberg2017unified}, using the default selection of \texttt{partition shap} for SHAP.

Among the gradient-based methods, we include \textbf{Gradient (Grad.)} \citep{simonyan2013deep}, \textbf{Integrated Gradients (Int. Grad.)} \citep{sundararajan2017axiomatic}, and their input-scaled variants: \textbf{Gradient $\times$ Input (Grad. $\times$ Input)} and \textbf{Integrated Gradients $\times$ Input (Int. Grad. $\times$ Input)}.

To evaluate \textit{faithfulness}, we compare our results using well-established erasure-based metrics: \textbf{comprehensiveness (C($\uparrow$))}, \textbf{sufficiency (S($\downarrow$))} \citep{deyoung2020eraser}, and \textbf{correlation with Leave-One-Out (L($\uparrow$))} scores \citep{jain2019attention}.

For \textit{plausibility} analysis, we followed DeYoung et al. \citep{deyoung2020eraser} and report \textbf{token-level Intersection-Over-Union (I($\uparrow$))}, \textbf{token-level F1 scores (F($\uparrow$))}, and \textbf{Area Under the Precision-Recall Curve (A($\uparrow$))}. For all measures, `$\uparrow$' signifies the higher the better and `$\downarrow$' signifies the lower the better.

We conduct all experiments using the \texttt{ferret-xai} library with its default (hyper)parameter settings \citep{attanasio2023ferret} and report the mean values for each measure.

\section{Experimental Results}

\subsection{Model Training \& Performance Analysis}
We evaluate five state-of-the-art transformer models that support Telugu: mBERT \citep{devlin2019bert}, XLM-R \citep{conneau2019unsupervised}, MuRIL \citep{khanuja2021muril}, IndicBERT \citep{joshi2022l3cube} and TeBERT \citep{joshi2022l3cube}. Detailed description of the models can be found in Appendix~\ref{Pre-trained}. Each model is trained in two settings: first, without rationales (denoted as \textit{wo/r}), and then with rationales (denoted as \textit{w/r}). Training a model without rationales allows it to learn any arbitrary class of functions, as it simply tries to fit the training data without any guidance on how its decisions align with human reasoning. In contrast, training with rationales explicitly curates the learning process by incorporating human-provided reasoning into the training. This allows the model to better align with human values during training. For the same, we adopt a specialized loss function, following the approach proposed by Zhang et al. \citep{zhong2019fine} and Pruthi et al. \citep{pruthi2022evaluating}: we took the mean of the rationale vector, added a small constant ($1\mathrm{e}{-8}$) in the place of zeroes ($0$), and converted it into a single probability distribution ($\hat{A}$) using softmax. Next, we computed the attention from the \texttt{[CLS]} token to other tokens in the last layer, averaged across all attention heads ($A$). The new modified loss function for this training is\footnote{For neutral sentences with no rationale, we only compute the CE loss.}: $
\text{Modified\ Loss} = \text{CE Loss} + \lambda \cdot \text{KL}(A \parallel \hat{A})$.

Across models, training with rationale supervision generally improves performance compared to label-only supervision. Four of the five architectures (mBERT, IndicBERT, MuRIL, and Te-BERT) show higher F1 and AUROC, with the largest relative gains observed in MuRIL and Te-BERT; accuracy in these models also increases. IndicBERT shows smaller but consistent improvements, while XLM-R is the only model for which performance remains comparable or slightly lower across settings. Thus, in relation to RQ1, \textit{these results indicate that incorporating rationales tends to enhance overall predictive performance rather than only changing accuracy alone. This trend has been largely consistent across contemporary text classification literature, which is primarily English-centric \citep{gao2024going, pruthi2022evaluating, herrewijnen2024human}, and to the best of our knowledge, this study is the first to investigate it for Telugu.} We refer readers to Table \ref {tab:model-performance} for a consolidated overview of the models' performance. For a detailed overview of the selection of hyperparameters for both types of training, we refer readers to Appendix~\ref{Hyp_Tun}.

\begin{table}[htbp]
\centering
\setlength{\tabcolsep}{4pt}
\begin{tabular}{l l|c|c|c}
\textbf{Model} & & \textbf{Accuracy} & \textbf{F1} & \textbf{AUROC} \\
\hline
\textbf{mBERT} & \textit{wo/r} & 0.664 & 0.655 & 0.820 \\

               & \textit{w/r} & 0.666 & 0.662 & 0.830 \\
\hline
\textbf{IndicBERT} & \textit{wo/r} & 0.713 & 0.711 & 0.879 \\

                   & \textit{w/r} & 0.715 & 0.718 & 0.886 \\
\hline
\textbf{MuRIL} & \textit{wo/r} & 0.698 & 0.699 & 0.865 \\

              & \textit{w/r}  & 0.712 & 0.715 & 0.873 \\
\hline
\textbf{XLM-R} & \textit{wo/r} & 0.705 & 0.705 & 0.864 \\

                & \textit{w/r}  & 0.692 & 0.695 & 0.867 \\
\hline
\textbf{Te-BERT} & \textit{wo/r} & 0.716 & 0.712 & 0.876 \\

              & \textit{w/r}  & 0.729 & 0.730 & 0.884 \\
\end{tabular}
\caption{Models' performance for both types of training. \textit{wo/r}: without rational, \textit{w/r}: with rational}
\label{tab:model-performance}

\end{table}

\subsection{\emph{Aligning} With Human Understanding}

Given the low-resource nature of Telugu, we do not attempt mechanistic or latent-space analyses (e.g., neuron activations or representational geometry), which are difficult to interpret without strong linguistic priors and diagnostic tools. Instead, we evaluate alignment through post-hoc explanations, which provide a human-interpretable view of model behavior. Our dataset includes explicit rationales collected from multiple annotators for every instance, offering a grounded representation of human reasoning. Importantly, post-hoc explanation methods are non-trainable and applied independently of the training procedure, ensuring that comparisons between label-only and rationale-supervised models are not confounded. We use plausibility metrics \citep{deyoung2020eraser} as a reasonable proxy for alignment, which measure how well model-generated explanations overlap with human-provided rationales. When explanations are reasonably faithful (i.e., explanations do reflect the underlying reasoning of the model \citep{jacovi2020towards}), higher plausibility indicates closer agreement between what the model attends to and what humans consider relevant--making plausibility a practical, interpretable proxy for human alignment in this setting.

Across models and explanation methods, rationale supervision yields consistent gains in plausibility, but not in faithfulness. For each plausibility (\textbf{A, F, I}) and faithfulness (\textbf{C, S, L}) metric, we perform paired two-sided t-tests and Wilcoxon signed-rank tests with a significance threshold of $p < 0.05$. We observe statistically significant improvements for \textbf{F} and \textbf{I} under both tests, while \textbf{A} shows a significant effect under Wilcoxon and a marginal trend under the t-test. In contrast, none of the faithfulness metrics are statistically significant under either test, indicating no systematic change in faithfulness due to rationale supervision. This divergence is expected. Faithfulness evaluates how accurately an explanation reflects a model’s internal reasoning process, whereas rationale supervision modifies the model itself rather than the explainer. Since post-hoc explainers are fixed, non-trainable tools \citep{rudin2019stop}, there is no inherent reason to expect systematic changes in faithfulness. Plausibility, by contrast, directly measures agreement with human-identified evidence. The observed improvements, therefore, indicate that rational supervision encourages models to rely on input regions that humans themselves deem relevant for sentiment decisions, making model behavior more aligned with human understanding. This directly addresses \textbf{RQ2}. To the best of our knowledge, contemporary literature lacks a systematic analysis of plausibility and faithfulness that is curated under a comparable experimental setup to ours, even in English-centric domains, let alone in low-resource settings. The work closest in spirit is by Carton et al. \citep{carton2022learn}, which reports marginal and variable improvements using a task-specific proxy faithfulness measure that may not be directly applicable to other settings, including ours. More broadly, existing studies in this area are fragmented and narrowly focused, making it difficult to draw transferable insights for our research question \citep{gao2024going}. Our detailed results are provided in Appendix \ref{DEL-REL}. We also report a complementary qualitative analysis with a few \textit{suitable} examples in Appendix \ref{QualiS} to show how rationale supervision can shifts token importance, without changing the final label, toward more \textit{plausible} task-relevant signals.

\subsection{Bias \& Fairness Analysis}
We analyze how alignment-oriented training interacts with social bias along gender and religion dimensions, where males and Hindus are treated as privileged groups based on the demographic distribution of native Telugu speakers. Overall, training with rationales induces only small relative changes in fairness metrics, including polarity distribution shift ($PDS$), true positive rate difference ($TPR_{diff}$), and false positive rate difference ($FPR_{diff}$), and these changes are not consistent across models or demographic groups. MuRIL and mBERT present notable exceptions in the case of religion bias, with a substantial decrease observed for MuRIL and a moderate increase for mBERT. Detailed fairness results are reported in Appendix~\ref{DEL-REL}. 

Answering \textbf{RQ3}, these results indicate that \textit{while alignment-oriented design choices reliably improve human alignment, they do not induce systematic or guaranteed improvements in fairness: models that become more aligned with human reasoning are not necessarily more fair. Though there's some occasional improvement in bias reduction in our findings, that is evidently not a clear indication that alignment-oriented supervision alone is sufficient to ensure consistent fairness improvements. This outcome is expected, as our training objective explicitly targets alignment with human understanding and does not impose fairness-specific constraints. Moreover, Telugu is an extremely low-resource language, and the pre-trained models considered here are trained on substantially less and less diverse data than those for high-resource languages. As a result, the scale and nature of social biases in Telugu models may differ from those reported in English-centric studies such as HateXplain}~\citep{mathew2021hatexplain}.

\section{Conclusion and Future Direction}
In this work, we argue that progress in low-resource sentiment analysis should be evaluated not only by accuracy, but also by alignment with human reasoning and context. We introduce \textbf{TeSent}, a Telugu sentiment dataset with human rationales, and \textbf{TeEEC}, a corpus for systematic bias evaluation. Using these, we show that rationale supervision improves alignment with human explanations and often enhances performance, while not necessarily leading to consistent fairness gains. Together, these resources provide a foundation for studying performance, alignment, and fairness in low-resource settings. As future work, we aim to mechanistically understand how rationale-based training influences learning dynamics and fairness in Telugu.

\section*{Limitation}
This work focuses on encoder-based transformer models and does not include experiments with large generative LLMs. To our knowledge, there is no well-established baseline for such models on Telugu sentiment analysis, particularly under the alignment-oriented setting we consider. Our analysis relies on human rationales and post-hoc explanation methods, which capture alignment signals but do not reveal model internals. While we introduce \textbf{TeEEC} to study fairness across gender and religion, we do not directly optimize fairness, and broader, intersectional biases remain beyond the scope of this work.

\section*{Funding Information \& Acknowledgement}
We are thankful to all the annotators and reviewers for their effort and valuable feedback. This work was supported by the \textit{UGPG Seed Grant Scheme} of SRM University AP, India, under Grant No. \texttt{SRMAP/UG-PG/SEED/2024-25/30}.

\section*{Dataset Release and Governance}
Annotator consent and institutional approval have been already obtained to support for this work. The dataset is released under a research-only EULA [\href{https://forms.gle/BM84eeBw8up6FLNN6}{LINK}] to ensure, to the best of the authors’ ability, non-commercial academic use, with restrictions on redistribution and commercial use and guidelines for responsible data handling specified in it.

\bibliography{ref}

\section*{Appendix}
\appendix
\section{Detailed Related Works Section}
\label{DRS}
\textbf{Sentiment classification benchmarks in English}\\
Initial resources for sentiment classification in English language comprises of product reviews~\citep{pang2002thumbs,maas2011learning}, and the relevance of sentiment classification of product reviews will probably never diminish. However, extensive use of social media platforms has drawn attention to understanding sentiment polarity of short social media posts~\citep{socher2013recursive,dong2014adaptive,wang2010latent,rosenthal2017semeval,orbach2021yaso}. Casual nature of such texts makes them extremely noisy containing grammatical errors, typos, an abundance of non-standard text, slang and code-mixing. This posed new challenges in curating benchmark datasets and addressing the problem itself. Mostly, sentiment classification has been considered as a binary~\citep{socher2013recursive,orbach2021yaso,pang2002thumbs,maas2011learning}, 3-class~\citep{dong2014adaptive}, or 5-class~\citep{wang2010latent,rosenthal2017semeval} problem. While most of the popular datasets sourced the data from one particular social media platform such as Twitter~\citep{dong2014adaptive}, IMDB~\citep{maas2011learning}, tripAdvisor~\citep{wang2010latent}, etc., datasets like~\citep{orbach2021yaso} considered diversifying sources over multiple platforms.

\noindent \textbf{Sentiment classification benchmarks in Indic languages}\\
The lack of high-quality datasets for sentiment classification in low-resource Indic languages has significantly limited research in this area. The most commonly cited corpora include Hindi~\citep{akhtar2016hybrid}, Bangla~\citep{islam2023sentigold}, Marathi~\citep{kulkarni2021l3cubemahasent}, multiple languages that include Telugu~\citep{doddapaneni2023towards} and code-mixed texts with English~\citep{patwa2020semeval,chakravarthi2020sentiment,chakravarthi2020corpus}. \citep{mukku2017actsa} and~\citep{marreddy2022resource} are the only popular Telugu sentiment corpus. Among the aforementioned Indic corpora, only Islam et al.~\citep{islam2023sentigold} follow a well-defined annotation protocol, described in detail, and have curated a large-scale dataset from Bangla texts across diverse domains, collected from multiple social media platforms. The remaining studies did not diversify their data sources. A few of them have small sample sizes~\citep{mukku2017actsa,chakravarthi2020sentiment}, some do not discuss the annotation protocol in detail~\citep{marreddy2022resource,kulkarni2021l3cubemahasent,akhtar2016hybrid}, and others suffer from weak data and annotations~\citep{doddapaneni2023towards,kulkarni2021l3cubemahasent,patwa2020semeval,chakravarthi2020sentiment,mukku2017actsa}. None of these works including Islam et al.~\citep{islam2023sentigold} extend their benchmarks to address explainability and fairness considerations. To the best of our knowledge, ours is the first large-scale Telugu sentiment corpus--and possibly the second among Indic languages after Islam et al.~\citep{islam2023sentigold}--comprising texts from diverse domains, sourced from multiple social media platforms, and developed using a well-defined and well-documented annotation protocol. Furthermore, it is the first Indic sentiment corpus to incorporate considerations of explainability and fairness.\\
\noindent \textbf{Explainability: }
Several of-the-shelf explainers have gained popularity for model auditing which are post-hoc in nature~\citep{bhatt2020explainable}. These explainers weight input tokens based on their contribution to the prediction. Since they are based on crude heuristics, these explainers are inherently imperfect \citep{rudin2019stop}. Plausibility and Faithfulness are the two primary categories of evaluation measures for these explainers~\citep{lyu2024towards}. A salient feature of our dataset is the inclusion of human-annotated rationales, which serve as ground truth for plausibility evaluation. While this has been explored to some extent in English~\citep{herrewijnen2024human}, our work is the first to address it in the context of an Indic language.
\noindent
\\
\textbf{Fairness} in sentiment classification has been evaluated in English for various social biases, including gender bias~\citep{thelwall2018gender,kiritchenko2018examining}, age-related bias~\citep{diaz2018addressing}, and racial bias~\citep{kiritchenko2018examining} among others. Attempts to evaluate fairness in Indic sentiment classification are rare, with only a few recent studies--such as those in Bangla~\citep{das2023toward,das2024colonial}--curating dataset for addressing this issue. Our work is the first attempt to provide corpus and subsequently evaluate fairness in Telugu sentiment classification, and a rare effort for the same across Indic languages.

\section{Preprocessing Pipeline}\label{App:Preprocessing}
This section details the preprocessing steps applied to the raw scraped corpus prior to annotation. To filter out non-Telugu content and code-mixed sentences from the collected corpus of 154,959 sentences, we utilized the UNICODE for Telugu and removed all sentences containing any non-Telugu fonts. However, we didn't remove punctuation, emojis, etc., and the size of the final collected data was roughly 120,000. From this, we removed sentences that had only emojis or punctuation, and we were left with roughly 90,000. From this collected data, we removed duplicates in two steps.

To address duplicates, we carried out a two-step deduplication process. First, we remove syntactically (near) duplicate sentences by computing the Jaccard similarity between bigrams of each sentence pair, retaining the first unique sentence, and discarding later ones that exceed a 75\% similarity threshold value. This step reduced the dataset to 86,519 unique sentences. In the second step, we used \href{https://huggingface.co/google-bert/bert-base-multilingual-cased}{mBERT} embeddings to remove semantically (near)duplicate sentences. For each new sentence, it calculates the pairwise cosine similarity between its embedding and the embeddings of all previously accepted, non-duplicate sentences. If the maximum similarity among these comparisons is below a 90\% threshold value, the sentence is considered sufficiently different and is kept, else discarded. After this semantic filtering, the dataset was reduced to 24,593 sentences. We next anonymized the dataset by replacing personally identifiable information such as email addresses, phone numbers, URLs, dates, etc, with applicable placeholders (\texttt{<NUMBER>, <EMAIL>, <LINK>, <PHONE>, <PERCENT>, <DATE>, <TIME>, <MONEY>}). After this step, we randomly selected 607 sentences for internal testing related to software validation and annotator recruitment, which left us with a working set of 23,986 sentences.

\section{Annotator Recruitment}\label{App:Recruitment}
This section describes the recruitment, screening, and compensation of annotators involved in the sentiment and rationale annotation task. First, we issued an open call through internal mailing lists, social media, and the internal organization networks of the organization, targeting native Telugu speakers. Prospective annotators (who are native speakers) were asked to complete a linguistic proficiency survey via Google Form. Participants labelled the sentiment and the rationales of a few selected Telugu sentences as Positive, Negative, or Neutral. Submissions were manually reviewed for accuracy, consistency, and quality of rationale selection by the Telugu-speaking authors against each sentiment. Out of 112 responses, based on the overall performance, a total of 95 annotators were selected by the Telugu-speaking authors for the main task. Next, a pilot annotation task was conducted among the Telugu-speaking authors to finalize the interface for our annotation software for web browsers for both mobile and desktop versions (such as whether ``double-click" or ``single-click" for selecting rationale would be more effective). We remunerated all annotators for completing each batch of 150 sentences with INR 100. However, among selected annotators, 21 defaulted without completing a single batch by the end of the annotation phase, leaving us a total of 74 active annotators.

\section{Annotation Software Design}\label{App:Software}
This section details the design principles and features of the custom-built annotation interface used in this study. The custom-built annotation interface included several features tailored to our design principles. Annotators logged in securely using unique credentials and were presented with one sentence at a time. For each sentence, they were required to select a primary sentiment label: Positive, Negative, or Neutral, and had the option to indicate an alternative sentiment, if applicable. A validation check is flagged in case of identical primary and alternative labels. Our interface is also equipped with an \textit{invalid flag} to remove invalid sentences on the fly, as described in section \ref{DCP}.

On the interface, rationale selection is done by double-clicking the word(s) from the sentence. These rationales, along with annotations, were stored in a \texttt{MongoDB} database. To ensure consistency and minimize fatigue, sentences were organized into batches of 150 sentences, with 10 batches forming one pool; in total, we used 17 such pools. Upon joining or completing a batch, annotators were assigned a random batch from the current pool, prioritizing batches with two existing annotations over those with one. This pool–batch structure ensured that every sentence received annotations from three independent annotators before moving to the next pool, mitigating risks from potential dropouts. This way, the annotation workload was distributed systematically to ensure reliability and reduce annotator fatigue. A progress bar, gamified leaderboard, and periodic manual reminders were used to maintain annotator engagement and ensure timely batch completion. A snapshot of the annotation interface is attached in Appendix \ref{AAI}.

\section{Descriptions of the Fairness Metrics}\label{F_MET}
\subsection{Counterfactual Test}
Counterfactual sentence pairs consist of two sentences, each drawn from the respective sets corresponding to the two demographic groups. Each counterfactual sentence represents one of the three sentiment classes $\in \text{\{negative, positive, neutral\}}$. Counterfactual pairs for the neutral class are generated using type (b) sentences by replacing the <person> placeholder with names and noun phrases taken from the name corpora of the two demographic groups. Since there is no one-to-one correspondence between names across demographic groups, counterfactual pairs are formed by considering all possible combinations of names, with one name drawn from each group. The 10 noun phrases exhibit a one-to-one correspondence across demographic groups, resulting in 10 counterfactual pairs for each type (b) template. Counterfactual pairs for the negative and positive class are generated using the type (a) sentences. For both the positive and negative classes, counterfactual pairs are constructed by replacing the <person> placeholder with names and noun phrases from the two demographic groups, for each <emotion word> corresponding to the respective sentiment polarity, as previously described. Given the large number of possible counterfactual pairs, we employ a randomized sampling algorithm to select 1320 counterfactuals for the case of gender, and 1200 counterfactuals for the case of religion, balanced across the three sentiment classes. The reported results are averaged over ten runs using different random seeds. The sampling algorithm is detailed in Appendix~\ref{Rand_Alg}. The counterfactual test measures the Polarity Difference Score (PDS), described as follows.

\noindent\textbf{Polarity Difference Score (PDS)}\\
\noindent PDS aims to answer the question: \textit {Does the classifier tend to assign more positive sentiment to the privileged demographic group compared to the underprivileged group?} Say, $\{(C_1^a,C_1^b),(C_2^a,C_2^b),...,(C_n^a,C_n^b)\}$ be the counterfactual pairs, where $a$ and $b$ represent the privileged and the under-privileged group respectively; the PDS is given as:
\begin{equation*}
     PDS = \frac{1}{n} \sum_{i=1}^n f(C_i^a) - f(C_i^b),
\end{equation*}
\noindent where $f(\cdot)$ gives the model predicted label mapped to an ordinal scale of sentiment polarity: $f(\text{negative})=1$, $f(\text{neutral})=2$, and $f(\text{positive})=3$.
\subsection{Equalized Odds (EO)}
A classifier satisfies EO if the privileged and the under-privileged group have equal $TPR$ (true positive rate) and $FPR$ (false negative rate). We precisely measure EO as the absolute difference of the TPR and FPR values, and denote them as $TPR_{diff}$ and $FPR_{diff}$. We consider the generated counterfactuals as the test set. As we work with a three-class problem, we consider the macro-average over the TPRs and FPRs of the three classes as the final TPR and FPR values.

\section{Detailed Result}
\label{DEL-REL}
Table~\ref{tab:bias_eval_g} and Table~\ref{tab:bias_eval_r} provide the results for gender and religion bias evaluation respectively.
Table~\ref{CombinedResults} provides Plausibility and Faithfulness evaluation Results.

\begin{table}[ht]
\centering
\renewcommand{\arraystretch}{1}

\resizebox{0.3\textwidth}{!}{%
\begin{tabular}{lc|ccc}
\textbf{Model} & & $PDS$ & $TPR_{diff}$ & $FPR_{diff}$ \\
\hline
\multirow{2}{*}{\textbf{mBERT}}
& \textit{wo/r} & -0.004 & 0.022 & 0.010 \\
 & \textit{w/r} & -0.004 & 0.025 & 0.013 \\
\hline
\multirow{2}{*}{\textbf{IndicBERT}}
 & \textit{wo/r}  & 0.003 & 0.006 & 0.003 \\
 & \textit{w/r} & 0.003 & 0.003 & 0.003 \\
\hline
\multirow{2}{*}{\textbf{MuRIL}}
 & \textit{wo/r} & 0.008 & 0.011 & 0.006 \\
 & \textit{w/r}  & 0.008 & 0.010 & 0.004 \\
\hline
\multirow{2}{*}{\textbf{XLM-R}}
 & \textit{wo/r}  & -0.004 & 0.017 & 0.007 \\
 & \textit{w/r}   & -0.007 & 0.008 & 0.004 \\
\hline
\multirow{2}{*}{\textbf{Te-BERT}}
 & \textit{wo/r} & 0.005 & 0.005 & 0.004 \\
 & \textit{w/r}  & -0.002 & 0.005 & 0.004 \\
\hline
\end{tabular}
}
\caption{Gender bias evaluation results. \textit{wo/r}: without rational, \textit{w/r}: with rational}
\label{tab:bias_eval_g}
\end{table}

\begin{table}[ht]
\centering
\renewcommand{\arraystretch}{1}

\resizebox{0.45\textwidth}{!}{%
\begin{tabular}{lc|ccc|ccc}
\textbf{Model} & & \multicolumn{3}{c|}{Hindu-Christian} & \multicolumn{3}{c}{Hindu-Muslim} \\
 & & $PDS$ & $TPR_{diff}$ & $FPR_{diff}$ & $PDS$ & $TPR_{diff}$ & $FPR_{diff}$ \\
\hline
\multirow{2}{*}{\textbf{mBERT}}
 & \textit{wo/r} & 0.023 & 0.012 & 0.014 & 0.045 & 0.025 & 0.029 \\
 & \textit{w/r} & 0.035 & 0.022 & 0.016 & 0.063 & 0.034 & 0.033  \\
\hline
\multirow{2}{*}{\textbf{IndicBERT}}
 & \textit{wo/r} & 0.004 & 0.019 & 0.013 & 0.012 & 0.011 & 0.004 \\
 & \textit{w/r} & 0.020 & 0.013 & 0.008 & 0.012 & 0.007 & 0.004\\
\hline
\multirow{2}{*}{\textbf{MuRIL}}
 & \textit{wo/r} & 0.048 & 0.033 & 0.020 & 0.051 & 0.036 & 0.020 \\
 & \textit{w/r}  &0.017 & 0.012 & 0.006 & 0.014 & 0.013 & 0.006 \\
\hline
\multirow{2}{*}{\textbf{XLM-R}}
 & \textit{wo/r} & 0.004 & 0.015 & 0.007 & 0.012 & 0.010 & 0.006 \\
 & \textit{w/r}  & 0.014 & 0.012 & 0.005 & 0.012 & 0.011 & 0.005 \\
\hline
\multirow{2}{*}{\textbf{Te-BERT}}
 & \textit{wo/r} & 0.003 & 0.005 & 0.004 & 0.010 & 0.006 & 0.006 \\
 & \textit{w/r} & 0.002 & 0.006 & 0.003 & 0.020 & 0.013 & 0.006 \\
\hline
\end{tabular}
}
\caption{Religion bias evaluation results. \textit{wo/r}: without rational, \textit{w/r}: with rational}
\label{tab:bias_eval_r}
\end{table}

\begin{table*}[ht]
\centering
\renewcommand{\arraystretch}{1.1}
\setlength{\tabcolsep}{3pt}
\resizebox{\textwidth}{!}{%
\begin{tabular}{c|lc|ccc|ccc|ccc|ccc|ccc|ccc}
\textbf{\rotatebox[origin=c]{90}{}} & \textbf{Model} & & \multicolumn{3}{c|}{\textbf{SHAP}} & \multicolumn{3}{c|}{\textbf{LIME}} & \multicolumn{3}{c|}{\textbf{Grad}} & \multicolumn{3}{c|}{\textbf{Grad $\times$ Input}} & \multicolumn{3}{c|}{\textbf{Int. Grad}} & \multicolumn{3}{c}{\textbf{Int. Grad $\times$ Input}} \\
\hline
 & & & A ($\uparrow$) & F ($\uparrow$) & I ($\uparrow$) & A ($\uparrow$) & F ($\uparrow$) & I ($\uparrow$) & A ($\uparrow$) & F ($\uparrow$) & I ($\uparrow$) & A ($\uparrow$) & F ($\uparrow$) & I ($\uparrow$) & A ($\uparrow$) & F ($\uparrow$) & I ($\uparrow$) & A ($\uparrow$) & F ($\uparrow$) & I ($\uparrow$) \\

\multirow{10}{*}{\rotatebox[origin=c]{90}{\textsc{Plausibility}}} 
 & \multirow{2}{*}{\textbf{mBERT}} & \textit{wo/r} & 0.568 & 0.396 & 0.277 & 0.565 & 0.392 & 0.271 & 0.549 & 0.370 & 0.257 & 0.409 & 0.262 & 0.167 & 0.422 & 0.281 & 0.180 & 0.522 & 0.363 & 0.247 \\
 
 & & \textit{w/r} & 0.570& 0.396& 0.278& 0.572& 0.405& 0.281& 0.556& 0.386& 0.270& 0.403& 0.257& 0.164& 0.425& 0.277& 0.176& 0.507& 0.348& 0.232\\
 & \multirow{2}{*}{\textbf{IndicBERT}} & \textit{wo/r} & 0.555& 0.419& 0.294& 0.549& 0.403& 0.277& 0.548& 0.444& 0.317& 0.392& 0.308& 0.208& 0.410& 0.315& 0.214& 0.572& 0.433& 0.308\\

 & & \textit{w/r} & 0.576& 0.430& 0.305& 0.562& 0.414& 0.288& 0.583& 0.463& 0.334& 0.416& 0.329& 0.225& 0.442& 0.329& 0.225& 0.573& 0.463& 0.298\\
 & \multirow{2}{*}{\textbf{MuRIL}} & \textit{wo/r} & 0.585& 0.446& 0.319& 0.594& 0.447& 0.318& 0.566& 0.436& 0.31& 0.424& 0.309& 0.207& 0.421& 0.295& 0.197& 0.584& 0.435& 0.31\\

 & & \textit{w/r} & 0.598& 0.448& 0.32& 0.607& 0.451& 0.321& 0.567& 0.447& 0.32& 0.415& 0.303& 0.202& 0.425& 0.311& 0.209& 0.6& 0.449& 0.32
\\
 & \multirow{2}{*}{\textbf{XLM-R}} & \textit{wo/r} & 0.577& 0.43& 0.303& 0.588& 0.438& 0.307& 0.56& 0.424& 0.298& 0.407& 0.296& 0.195& 0.395& 0.271& 0.177& 0.508& 0.378& 0.259
\\

 & & \textit{w/r} & 0.575& 0.429& 0.303& 0.582& 0.432& 0.303& 0.554& 0.427& 0.303& 0.38& 0.275& 0.181& 0.395& 0.291& 0.191& 0.477& 0.367& 0.249
\\
 & \multirow{2}{*}{\textbf{Te-BERT}} & \textit{wo/r} & 0.578& 0.437& 0.309& 0.595& 0.448& 0.316& 0.551& 0.422& 0.298& 0.385& 0.277& 0.187& 0.418& 0.281& 0.186& 0.596& 0.445& 0.318
\\

 & & \textit{w/r} & 0.583& 0.437& 0.31& 0.604& 0.459& 0.327& 0.557& 0.423& 0.299& 0.386& 0.271& 0.182& 0.448& 0.323& 0.217& 0.614& 0.454& 0.329
\\
\hline
 & & & C ($\uparrow$) & S ($\downarrow$) & L ($\uparrow$) & C ($\uparrow$) & S ($\downarrow$) & L ($\uparrow$) & C ($\uparrow$) & S ($\downarrow$) & L ($\uparrow$) & C ($\uparrow$) & S ($\downarrow$) & L ($\uparrow$) & C ($\uparrow$) & S ($\downarrow$) & L ($\uparrow$) & C ($\uparrow$) & S ($\downarrow$) & L ($\uparrow$) \\
\multirow{10}{*}{\rotatebox[origin=c]{90}{\textsc{Faithfulness}}} 
 & \multirow{2}{*}{\textbf{mBERT}} & \textit{wo/r} & 0.426& -0.021& 0.239& 0.456& -0.036& 0.281& 0.267& 0.101& 0.062& 0.118& 0.275& -0.033& 0.147& 0.262& -0.01& 0.35& 0.059& 0.156
\\

 & & \textit{w/r} & 0.442& 0.008& 0.232& 0.479& -0.019& 0.273& 0.3& 0.12& 0.064& 0.121& 0.329& -0.05& 0.164& 0.304& -0.004& 0.342& 0.136& 0.123
\\
 & \multirow{2}{*}{\textbf{IndicBERT}} & \textit{wo/r} & 0.428& 0.061& 0.212& 0.505& 0.032& 0.359& 0.346& 0.16& 0.038& 0.209& 0.281& 0.019& 0.188& 0.301& 0.001& 0.397& 0.101& 0.146
\\

  & & \textit{w/r} & 0.479& 0.05& 0.261& 0.562& 0.029& 0.413& 0.376& 0.158& 0.055& 0.217& 0.282& 0.039& 0.199& 0.306& 0.025& 0.437& 0.108& 0.168
\\
 & \multirow{2}{*}{\textbf{MuRIL}} & \textit{wo/r} & 0.486& 0.036& 0.265& 0.505& 0.03& 0.307& 0.352& 0.119& 0.053& 0.117& 0.352& -0.073& 0.156& 0.316& -0.016& 0.461& 0.038& 0.217
\\

  & & \textit{w/r} & 0.454& 0.011& 0.253& 0.473& 0.007& 0.274& 0.334& 0.117& 0.044& 0.094& 0.34& -0.077& 0.125& 0.311& -0.015& 0.439& 0.02& 0.235
\\
 & \multirow{2}{*}{\textbf{XLM-R}} & \textit{wo/r} & 0.481& -0.014& 0.271& 0.512& -0.025& 0.35& 0.332& 0.097& 0.075& 0.095& 0.345& -0.115& 0.119& 0.294& -0.02& 0.364& 0.048& 0.155
\\

  & & \textit{w/r} & 0.481& -0.027& 0.299& 0.5& -0.029& 0.374& 0.309& 0.112& 0.073& 0.07& 0.359& -0.121& 0.135& 0.267& 0.01& 0.331& 0.09& 0.133
\\
 & \multirow{2}{*}{\textbf{Te-BERT}} & \textit{wo/r} & 0.455& 0.046& 0.26& 0.483& 0.035& 0.319& 0.326& 0.123& 0.04& 0.096& 0.341& -0.083& 0.113& 0.356& -0.035& 0.439& 0.041& 0.237
\\

  & & \textit{w/r} & 0.451& 0.026& 0.265& 0.476& 0.017& 0.319& 0.322& 0.118& 0.024& 0.097& 0.346& -0.062& 0.161& 0.267& 0.021& 0.417& 0.034& 0.212
\\
\end{tabular}
}
\caption{Combined Plausibility and Faithfulness Results. \textit{wo/r}: without rational, \textit{w/r}: with rational}
\label{CombinedResults}
\end{table*}

\section{Sampling Algorithm for Generating Counterfactuals}
\label{Rand_Alg}
\subsection{Gender Bias}
To assess gender bias, 1320 counterfactual pairs are generated. This
includes i) 440 neutral pairs created using the four templates from Type(b) with 100 random male-female named pairs stratified across
Hindu, Muslim, and Christian names, plus 10 noun-phrase pairs,
per template and ii) 440 positive and negative pairs each, stratified
across the three religions, using the seven templates from Type(a):
for each of 10 randomly chosen <emotion> word of a polarity,
generate 44 pairs by randomly selecting a template and a random
named pair or noun phrase pair.
\subsection{Religion Bias}
To generate Religion-based Bias Dataset, we adapt a similar sampling procedure as discussed for Gender-based Bias Dataset, but
discarding the noun phrases, as they do not contribute to any religion. We generate 1200 counterfactual pairs for each combination
of the three religions (Hindu, Muslim, and Christian). This includes,
i) 400 neutral pairs created using Type(b) templates with 100 random named pairs of a chosen religion combination stratified across
female and male groups, per template and ii) 400 positive and negative pairs each, stratified across the two gender groups, using the
seven templates from Type(a): for each 10 randomly chosen <emotion> word of a polarity, generate 40 pairs by randomly selecting a
template and a random named pair.

\section{Qualitative Analysis for Plausibility}\label{QualiS}
In this section, we have selected (a few) \textit{comparable} examples where both versions ({\textit{w/r}}, {\textit{wo/r}}) of a model predict the ground-truth label, and rank them by the difference in plausibility scores: $\mathrm{AUPRC}_{\textit{w/r}} -\mathrm{AUPRC}_{\textit{wo/r}}$. For each example, we report the Top-$K$ (here, $K = 4$) union of annotator-provided rationales and token importance scores computed via \textbf{SHAP} for both \textit{w/r} and \textit{wo/r} models. We then provide a brief observation highlighting the resulting shift in model behaviour in Table \ref{QualiSTable}. We use Google Translate (via web search) to obtain English translations, which are then reviewed by Telugu-speaking authors to ensure reasonable fidelity to the original text.

\begin{table*}[t]
\centering
\scriptsize
\setlength{\tabcolsep}{3pt}
\renewcommand{\arraystretch}{1.2}

\begin{tabular}{
>{\RaggedRight\arraybackslash}p{0.10\textwidth}
|>{\RaggedRight\arraybackslash}p{0.20\textwidth}
|>{\RaggedRight\arraybackslash}p{0.20\textwidth}
|>{\RaggedRight\arraybackslash}p{0.05\textwidth}
|>{\RaggedRight\arraybackslash}p{0.15\textwidth}
|>{\RaggedRight\arraybackslash}p{0.20\textwidth}
|>{\RaggedRight\arraybackslash}p{0.10\textwidth}
}

\textbf{Model} & \textbf{Telugu Input} & \textbf{English Input} & \textbf{Label} & \textbf{Rationales' Union)} & \textbf{Top Tokens (\textit{wo/r} vs.\ \textit{w/r})} & \textbf{Observation} \\

\hline

\textbf{mBERT}
&
\texttelugu{పోలీసు అధికారులు అతి చేస్తున్నారు రైతులు అలా చేయటంలో తప్పు లేదు}
&
Police officials are going too far... there is nothing wrong with the farmers acting that way.
&
\textbf{N}
&
\texttelugu{అతి} (excess), \texttelugu{తప్పు} (wrong), \texttelugu{చేస్తున్నారు} (doing), \texttelugu{లేదు} (not)
&
\textit{wo/r}: \texttelugu{లేదు} (not), \texttelugu{అతి} (excess) \newline
\textit{w/r}: \texttelugu{అతి} (excess), \texttelugu{చేస్తున్నారు} (doing)
&
\textit{w/r} model captures main negative trigger. \\

\textbf{IndicBERT}
&
\texttelugu{గతంతో పోలిస్తే సమయపాలన చాలా మెరుగ్గా ఉంటుంది}
&
Compared to the past, punctuality is much better.
&
\textbf{P}
&
\texttelugu{మెరుగ్గా} (better)
&
\textit{wo/r}: \texttelugu{చాలా} (very), \texttelugu{పోలిస్తే} (compared) \newline
\textit{w/r}: \texttelugu{మెరుగ్గా} (better), \texttelugu{సమయపాలన} (punctuality)
&
\textit{w/r} model emphasizes sentiment word. \\

\textbf{IndicBERT}
&
\texttelugu{ఈ ఊరు ఏ జిల్లా ఏ మండలం లో ఉంది ఈ గ్రామం పేరు ఏమిటి}
&
In which district and Mandal is this town located? What is the name of this village?
&
\textbf{Nt}
&
\texttelugu{ఏమిటి} (what)
&
\textit{wo/r}: \texttelugu{మండలం} (mandal), \texttelugu{పేరు} (name) \newline
\textit{w/r}: \texttelugu{ఏమిటి} (what), \texttelugu{పేరు} (name)
&
\textit{w/r} model aligns with interrogative intent. \\

\textbf{IndicBERT}
&
\texttelugu{సభ అనంతరం ఆంతరంగికులతో మహారాజు సమావేశమయ్యారు}
&
After the assembly, the Maharaja met with his inner circle.
&
\textbf{Nt}
&
\texttelugu{సమావేశమయ్యారు} (met), \texttelugu{మహారాజు} (Maharaja)
&
\textit{wo/r}: \texttelugu{అనంతరం} (after) \newline
\textit{w/r}: \texttelugu{సమావేశమయ్యారు} (met)
&
\textit{w/r} model focuses on action. \\

\textbf{MuRIL}
&
\texttelugu{రితిక లాంటి ఆమె ప్రతి ఉమ్మడి కుటుంబం లో ఒక్కరు ఉంటారు రితిక సహజ నటి}
&
There is one person like Ritika in every joint family. Ritika is a natural actress.
&
\textbf{P}
&
\texttelugu{సహజ} (natural), \texttelugu{నటి} (actress)
&
\textit{wo/r}: \texttelugu{ఆమె} (she), \texttelugu{ప్రతి} (every) \newline
\textit{w/r}: \texttelugu{సహజ} (natural), \texttelugu{నటి} (actress)
&
\textit{w/r} model highlights descriptive sentiment. \\

\textbf{MuRIL}
&
\texttelugu{నాది చిత్తూరు జిల్లా అన్న చాలా అంటే చాలా అభిమానం అన్న సిఎం}
&
I hail from Chittoor district, brother—and I have immense, truly immense affection for the CM.
&
\textbf{P}
&
\texttelugu{అభిమానం} (affection), \texttelugu{చాలా} (very)
&
\textit{wo/r}: \texttelugu{అన్న} (brother), \texttelugu{చాలా} (very) \newline
\textit{w/r}: \texttelugu{అభిమానం} (affection), \texttelugu{సిఎం} (CM)
&
\textit{w/r} model strengthens sentiment signal. \\

\textbf{Te-BERT}
&
\texttelugu{బీబీసీ పశ్చిమ గోదావరిని తూర్పు గోదావరిని కలిపే చించినాడ బ్రిడ్జి కూడా ఇదే పరిస్థితి ఐరన్ పైకి కనిపిస్తుంది తూర్పు పశ్చిమ గోదావరి రోడ్లైతే మరి దారుణం}
&
The Chinchinada Bridge—which connects West Godavari and East Godavari—is in this very same condition. The exposed iron reinforcement is clearly visible. As for the roads in East and West Godavari, their state is even more appalling.
&
\textbf{N}
&
\texttelugu{దారుణం} (appalling)
&
\textit{wo/r}: \texttelugu{మరి} (very), \texttelugu{దారుణం} (terrible) \newline
\textit{w/r}: \texttelugu{దారుణం} (appalling)
&
\textit{w/r} model isolates strong negative word. \\

\hline
\end{tabular}

\caption{Qualitative comparison}
\label{QualiSTable}
\end{table*}

\section{Pre-trained Models}
\label{Pre-trained}
\subsection{mBERT}
mBERT (BERT-base-multilingual-cased) is based on Google's BERT-base transformer model (with 12 Layers and $\sim 100$ million parameters) trained on Wikipedia texts in 104 languages, including Telugu for the Masked Language Modeling (MLM) and Next Sentence Prediction tasks~\citep{devlin2019bert}. Although not specifically trained on Telugu alone, mBERT has demonstrated strong performance in sentiment classification tasks due to its shared multilingual representation~\citep{hedderich2021survey}. For Telugu sentiment classification, mBERT supports cross-lingual transfer capabilities with acceptable performance even with limited data on Telugu. Its ability to generalize across languages makes it effective for multilingual applications\citep{kalyan2021ammus}, although it may not capture fine-grained Telugu-specific nuances as well as regionally tuned models. However, since it is not optimized for Telugu morphology or syntax, its performance may lag~\citep{wu2020all} behind Telugu-specialized models such as IndicBERT, L3Cube-Telugu-BERT, in capturing language-specific nuances. Nonetheless, mBERT remains a powerful yet reliable baseline that is widely used in academic research, particularly in low-resource settings~\citep{marreddy2022resource,rajalakshmi2023hottest,park2klue,marreddy2021clickbait, duggenpudi2022teluguner}.

\subsection{XLM-R}
XLM-RoBERTa (XLM-R) is a general-purpose multilingual transformer model developed by Facebook AI, specifically designed to improve cross-lingual understanding by training on a massive multilingual corpus (2.5TB of filtered ``Common Crawl'' data across 100+ languages, including Telugu) for MLM task (with no NSP as used in mBERT)~\citep{conneau2019unsupervised}. As it purely focuses on MLM tasks, it provides better contextual modeling and transfer learning than mBERT, having improved downstream performance~\citep{hedderich2021survey}. Therefore, XLM-R can lend good performance for Telugu sentiment analysis, particularly when fine-tuned with local data for benchmarking\citep{kulkarni2021l3cubemahasent,joshi2022l3cube}. However, Telugu-specific models like MuRIL or L3Cube-Telugu-BERT may offer better cultural and linguistic alignment~\citep{das2022hate, rajalakshmi2023hottest}.
\subsection{MuRIL}
MuRIL (Multilingual Representations for Indian Languages) is also a transformer-based BERT model, specifically designed to support 17+ Indian languages, including Telugu and English~\citep{khanuja2021muril}. Unlike mBERT, MuRIL is pre-trained on a large corpus of Telugu sentences from web, religious script, news data, etc., for the Masked Language Modeling (MLM) and Translation Language Modeling (TLM) tasks, offering better understanding of Telugu morphology and syntax~\citep{joshi2022l3cube}. As the pre-training data favors informal texts from the web, MuRIL can be less effective for formal and classical Telugu NLP tasks. However, MuRIL is a strong candidate for Telugu sentiment classification, especially when analyzing informal, social media, or conversational data, making it superior to general multilingual models like mBERT and XLM-R~\citep{das2022hate,rajalakshmi2023hottest}.
\subsection{IndicBERT}
IndicBERT (ai4bharat/indicBERTv2-MLM-only) is a multilingual BERT-like model developed by AI4Bharat, trained on OSCAR and AI4Bharat curated corpora of 12 Indian languages (including Telugu and English) for MLM task~\citep{joshi2022l3cube}. It is well-suited for monolingual Telugu NLP tasks rather than cross-lingual ones and do not support code-mixed data. For Telugu sentiment classification, it provides language-aware tokenization, clean embeddings, and faster training~\citep{marreddy2022resource,rajalakshmi2023hottest,duggenpudi2022teluguner}.
\subsection{Te-BERT} 
Telugu-BERT (L3Cube-Telugu-BERT) is a transformer-based BERT model pre-trained specifically on Telugu text (Telugu OSCAR, Wikipedia, news) by the L3Cube Pune research group for MLM task~\citep{joshi2022l3cube}. Since Telugu-BERT is tailored for Telugu, it excels in capturing the vocabulary, syntax, and semantics of the language, thus able to recognize nuanced expressions, idioms, and sentiments that are often poorly represented in multilingual models like mBERT and XLM-R. Therefore, it is ideal for researchers working on pure Telugu text analysis with sufficient labeled data for fine-tuning.

\section{Hyper-parameter Tuning}
\label{Hyp_Tun}
We split the dataset into 80\% training, 10\% validation, and 10\% test sets using stratified sampling. On the validation set,  we applied grid search hyperparameter tuning across all selected models. For each model, we explored a range of batch sizes (16, 32, 64) and learning rates (LR) from 1e-5 to 5e-5. For training with rationale, we varied the regularization parameter $\lambda$ from 0.1 to 0.7. After extensive tuning, the best performance was observed with $\lambda = 0.7$, which consistently delivered optimal results across all models. Finally, for both types of training, we used batch size as 64, 2e-5 as LR, and the Adam optimizer, and obtained consistently strong results with 4 epochs. We used a standard Python 3 Google Colab (Pro) runtime with one L4 GPU for all experiments, including model training and explanation generation, over a total duration of approximately 14 hours. 

\section{Annotation Interface}\label{AAI}
\begin{figure*}[t]
    \centering
    \includegraphics[width=0.9\textwidth]{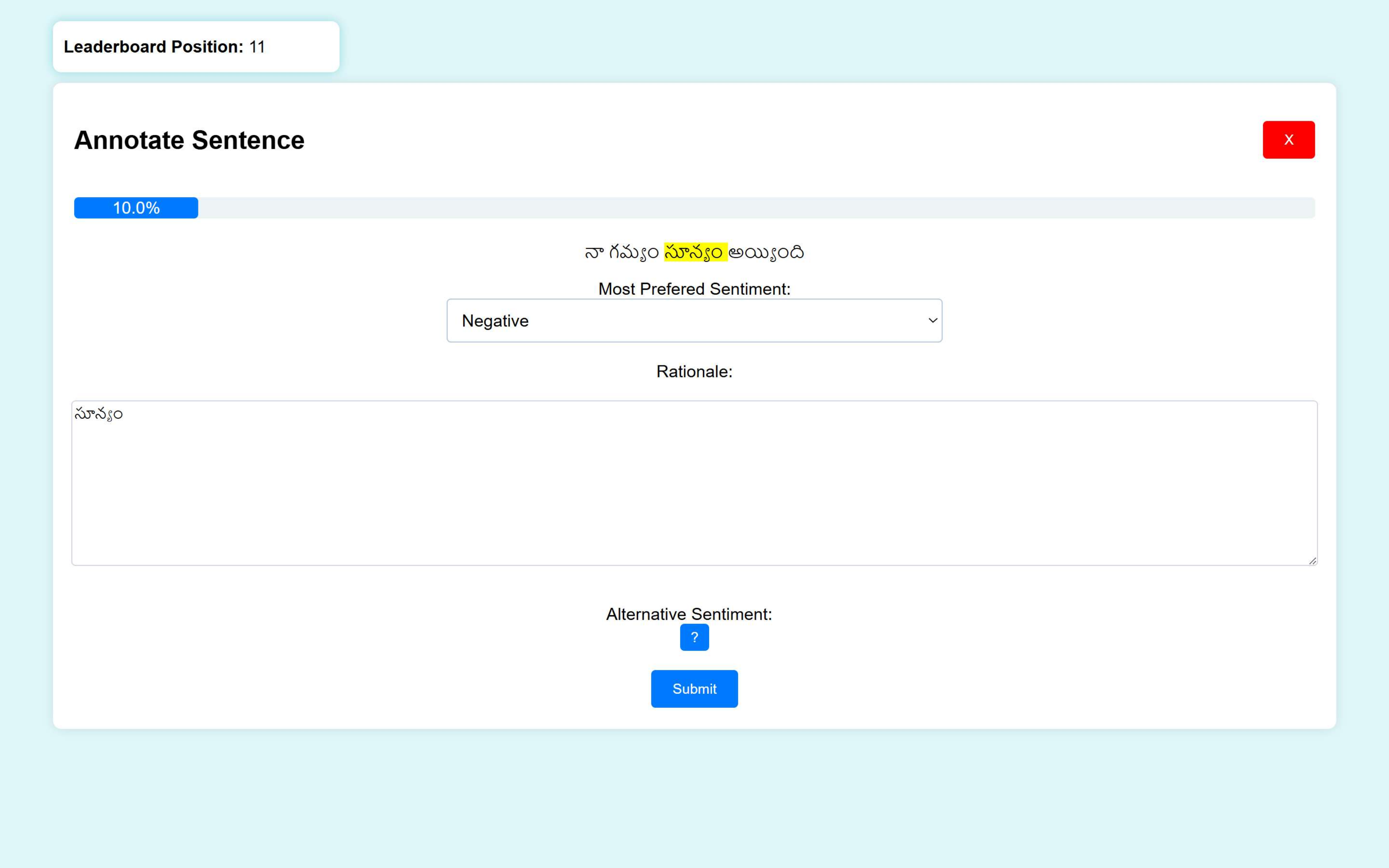}
    \caption{Annotation Interface}
    \label{fig:annotation_interface}
\end{figure*}

Figure \ref{fig:annotation_interface} presents a self-explanatory snapshot of our annotation interface.

\section{Telugu Topics Overview}\label{AC}

\begin{flushleft}
For brevity, we are listing the top 3 search phrases for each category, using which we collected $>90\%$ of the comments.

Religion: "Role of religion in Telugu society", "Telugu religious beliefs and practices", "Influence of Telugu faith in modern politics"\\
Education: "Telugu education system reform", "Online Telugu learning platforms", "Future of education in Telugu medium schools"\\
Personality: "Telugu personality development programs", "Analysis of Telugu personality traits", "Psychology of Telugu personalities"\\
Food \& Cooking: "Popular Telugu recipes and dishes", "Modern trends in Telugu cooking", "Healthy eating in Telugu cuisine"\\
Shopping: "Telugu online shopping behaviour", "Telugu consumer preferences and habits", "Black Friday deals in Telugu markets"\\
Crime: "Telugu crime rate analysis", "Notable Telugu true crime cases", "Telugu criminal justice system overview"\\
Law \& Justice: "Telugu legal system reform efforts", "Access to justice in Telugu communities", "High-profile Telugu court case summaries"\\
Trade: "Telugu international trade discussions", "Impact of trade wars on Telugu businesses", "Telugu trade policy analysis"\\
Entertainment: "Latest Telugu entertainment updates", "Trends in Telugu Tollywood industry", "Telugu movie reviews and ratings"\\
Sports: "Highlights of Telugu sports achievements", "Telugu athletes in Olympics history", "Debates in Telugu sports media"\\
Technology: "Innovations in Telugu tech startups", "Role of AI in Telugu applications", "Future of Telugu digital technologies"\\
People: "Inspiring Telugu personal stories", "Influential Telugu people in history", "Interviews with Telugu celebrities"\\
Science: "Scientific breakthroughs by Telugu researchers", "Latest Telugu science fair projects", "Space exploration in Telugu media"\\
Government: "Telugu government policy updates", "Public administration in Telugu regions", "Governance challenges in Telugu states"\\
Health: "Telugu health awareness campaigns", "Mental health in Telugu society", "Overview of Telugu healthcare systems"\\
Product: "Telugu tech product reviews", "Best Telugu consumer electronics", "Top Telugu products of 2024"\\
International: "International Telugu diaspora news", "Global events affecting Telugu communities", "Telugu perspectives on international relations"\\
Election: "Telugu election coverage highlights", "Voter turnout trends in Telugu states", "Analysis of Telugu election results"\\
Service: "Customer service trends in Telugu markets", "Best service industries for Telugu users", "Innovations in Telugu public service delivery"\\
Politics: "Telugu political debates and opinions", "Current affairs in Telugu politics", "Telugu party manifestos and analysis"\\
Job: "Telugu job market opportunities", "Career guidance for Telugu graduates", "Getting hired in Telugu job portals"\\
Social Media: "Telugu social media influencers", "Trends on Telugu social platforms", "Impact of social media on Telugu youth"\\
News: "Breaking Telugu news stories", "World news from Telugu perspective", "Top Telugu news highlights"\\
Environment: "Telugu discussions on climate change", "Sustainability efforts in Telugu regions", "Environmental issues in Telugu states"\\
Accident: "Telugu accident news reports", "Road safety tips for Telugu drivers", "Prevention of accidents in Telugu regions"\\
Transportation: "Public transport in Telugu cities", "Rise of electric vehicles in Telugu states", "Future of Telugu transportation systems"\\
Economics: "Telugu economic development trends", "State of Telugu regional economies", "Economic theory in Telugu discourse"\\
Stock Market: "Telugu stock market tips and strategies", "Investing trends among Telugu traders", "Telugu market news and updates"\\
Others: "Trending topics in Telugu discussions", "Popular Telugu YouTube content", "Miscellaneous Telugu news highlights"

\end{flushleft}

\end{document}